\documentclass[10pt,twocolumn,letterpaper]{article}

\PassOptionsToPackage{table}{xcolor}
\usepackage[pagenumbers]{cvpr}
\usepackage{multirow}
\usepackage{tabularx}

\definecolor{cvprblue}{rgb}{0.21,0.49,0.74}
\usepackage[pagebackref,breaklinks,colorlinks,allcolors=cvprblue]{hyperref}

\def\confName{CVPR}
\def\confYear{2027}

\title{Textual Response-Aware Guidance for Zero-Shot Adversarial Robustness of Vision-Language Models}

\author{Yubo Cui$^1$, Xianchao Guan$^1$, Zheng Zhang$^{1,2}$\\
$^1$Harbin Institute of Technology, Shenzhen, China; $^2$Shenzhen Loop Area Institute, Shenzhen, China\\
{\tt\small \{yubo{\_}cui, guanxianchao\}@stu.hit.edu.cn, darrenzz219@gmail.com}
}

\makeatletter
\patchcmd{\@maketitle}{\vspace*{24pt}}{\vspace*{16pt}}{}{}
\patchcmd{\@maketitle}{\vspace*{12pt}}{\vspace*{6pt}}{}{}
\makeatother

\begin{document}
\maketitle
\begin{abstract}
Vision-language models (VLMs) exhibit strong zero-shot generalization but remain vulnerable to adversarial attacks.
Existing adversarial fine-tuning methods often use pre-trained model guidance to preserve image--text alignment while improving robustness.
Our analysis of visual representation drift reveals complementary limitations of two representative forms of guidance.
Matching visual features provides broad coverage by constraining all directions equally, but lacks semantic emphasis.
Matching prediction distributions prioritizes semantically relevant directions, but remains limited by the candidate texts and prediction probabilities.
These limitations can hinder preservation of pre-trained image--text relationships and degrade zero-shot performance.
To address these issues, we propose \textbf{T}extual \textbf{R}esponse-\textbf{A}ware \textbf{G}uidance (\textbf{TRAG}), which constrains visual representation drift using the pre-trained text encoder's responses to local variations in input embeddings.
Specifically, TRAG constructs anchor-level constraints from the response directions and magnitudes around semantic anchors, then aggregates them using pre-trained image--anchor relevance, yielding an image-adaptive constraint.
Extensive experiments across architectures and tasks demonstrate state-of-the-art clean and robust performance.
Further analysis shows that TRAG helps preserve pre-trained image--text relationships.
\end{abstract}
\vspace{-12pt}
    
\section{Introduction}
\label{sec:intro}
\vspace{-3pt}

Vision-language models (VLMs)~\cite{CLIP,CoCa} achieve strong zero-shot generalization through large-scale pre-training on image--text pairs.
By aligning visual and textual representations in a shared embedding space, VLMs support applications such as image classification~\cite{CoOp} and image--text retrieval~\cite{BLIP}, and provide visual backbones for large vision-language models (LVLMs)~\cite{LLaVA}.
However, similar to conventional deep neural networks~\cite{SzegedyAdversarialExamples,GoodfellowAdversarialExamples}, VLMs remain vulnerable to adversarial examples.
Small, carefully crafted image perturbations can induce incorrect predictions, undermining model reliability~\cite{CoAttack,LVLMEvaluation}.
Adversarial fine-tuning improves robustness but can compromise zero-shot generalization~\cite{TeCoA}.
A central challenge is therefore to preserve the pre-trained image--text alignment underlying zero-shot transfer while improving adversarial robustness.

\newlength{\figonecontentwidth}
\begin{figure}[t]
  \centering
  \setlength{\figonecontentwidth}{\dimexpr\linewidth-2pt\relax}
  \begin{subfigure}[t]{0.320078\figonecontentwidth}
    \centering
    \includegraphics[width=\linewidth,trim={60bp 25bp 60bp 55bp},clip]{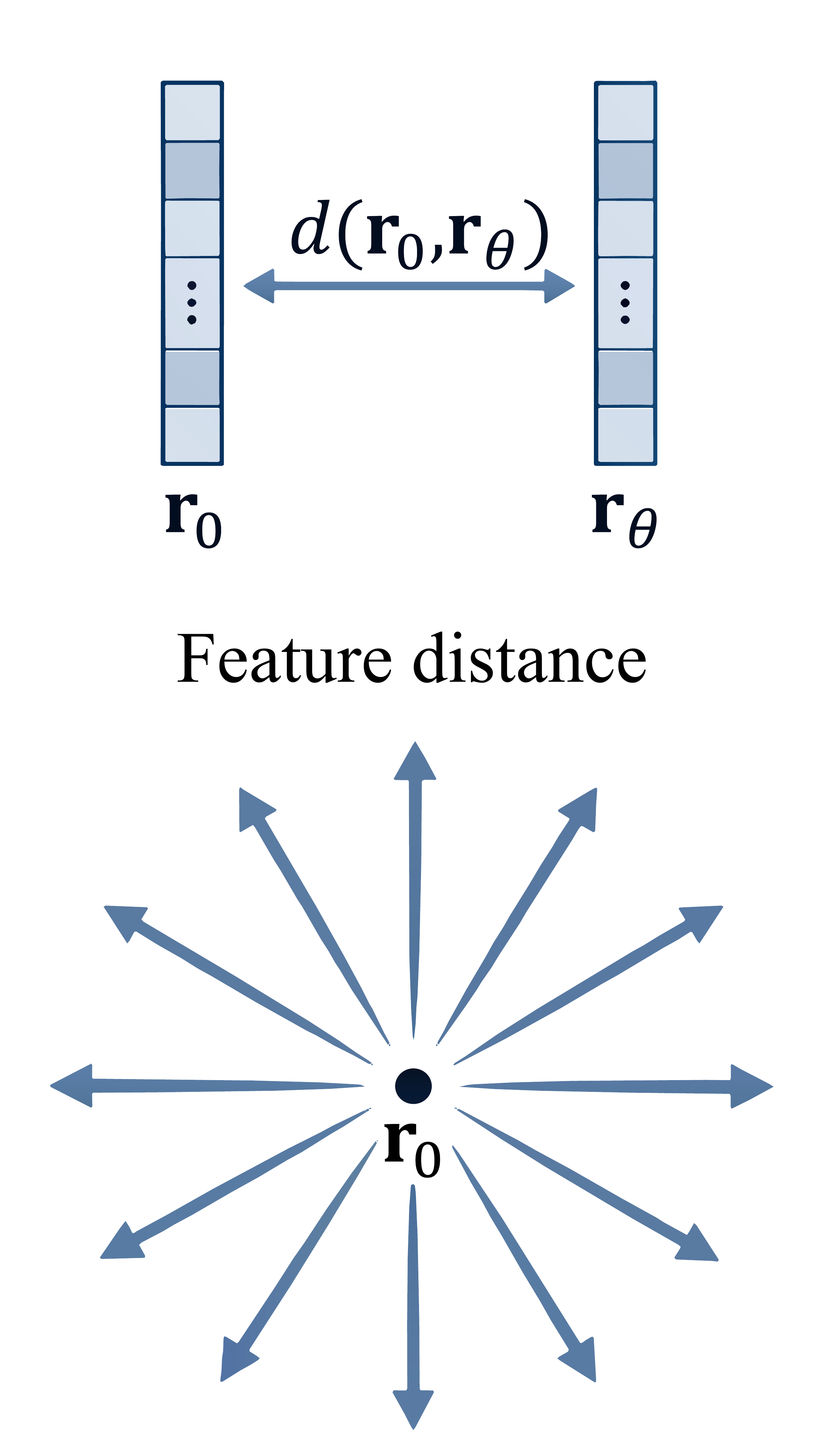}
    \caption{Feature matching}
    \label{fig:motivation-a}
  \end{subfigure}\hspace{1pt}\begin{subfigure}[t]{0.356523\figonecontentwidth}
    \centering
    \includegraphics[width=\linewidth,trim={0bp 25bp 0bp 55bp},clip]{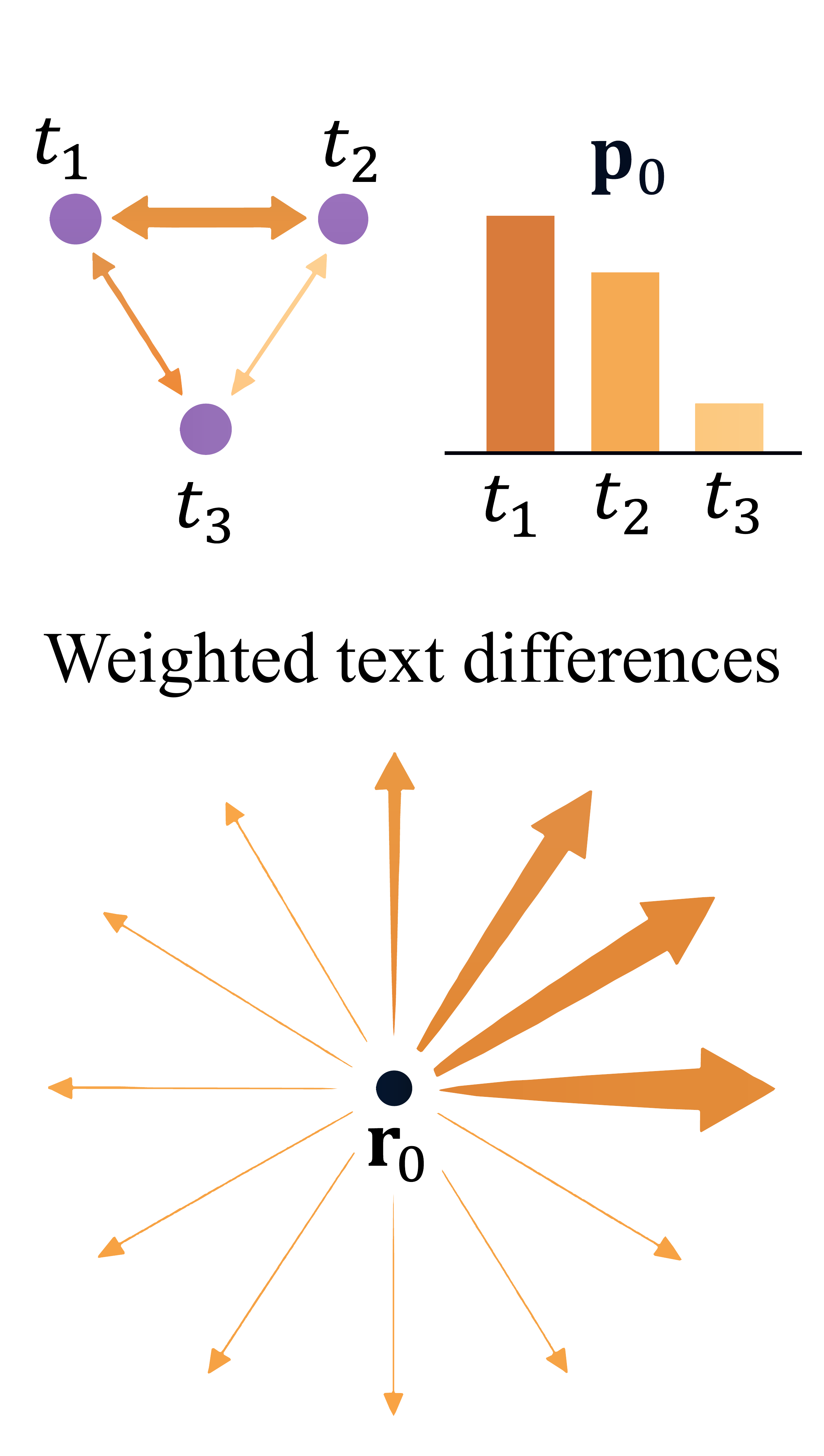}
    \caption{Distribution matching}
    \label{fig:motivation-b}
  \end{subfigure}\hspace{1pt}\begin{subfigure}[t]{0.323399\figonecontentwidth}
    \centering
    \includegraphics[width=\linewidth,trim={54.5bp 25bp 54.5bp 55bp},clip]{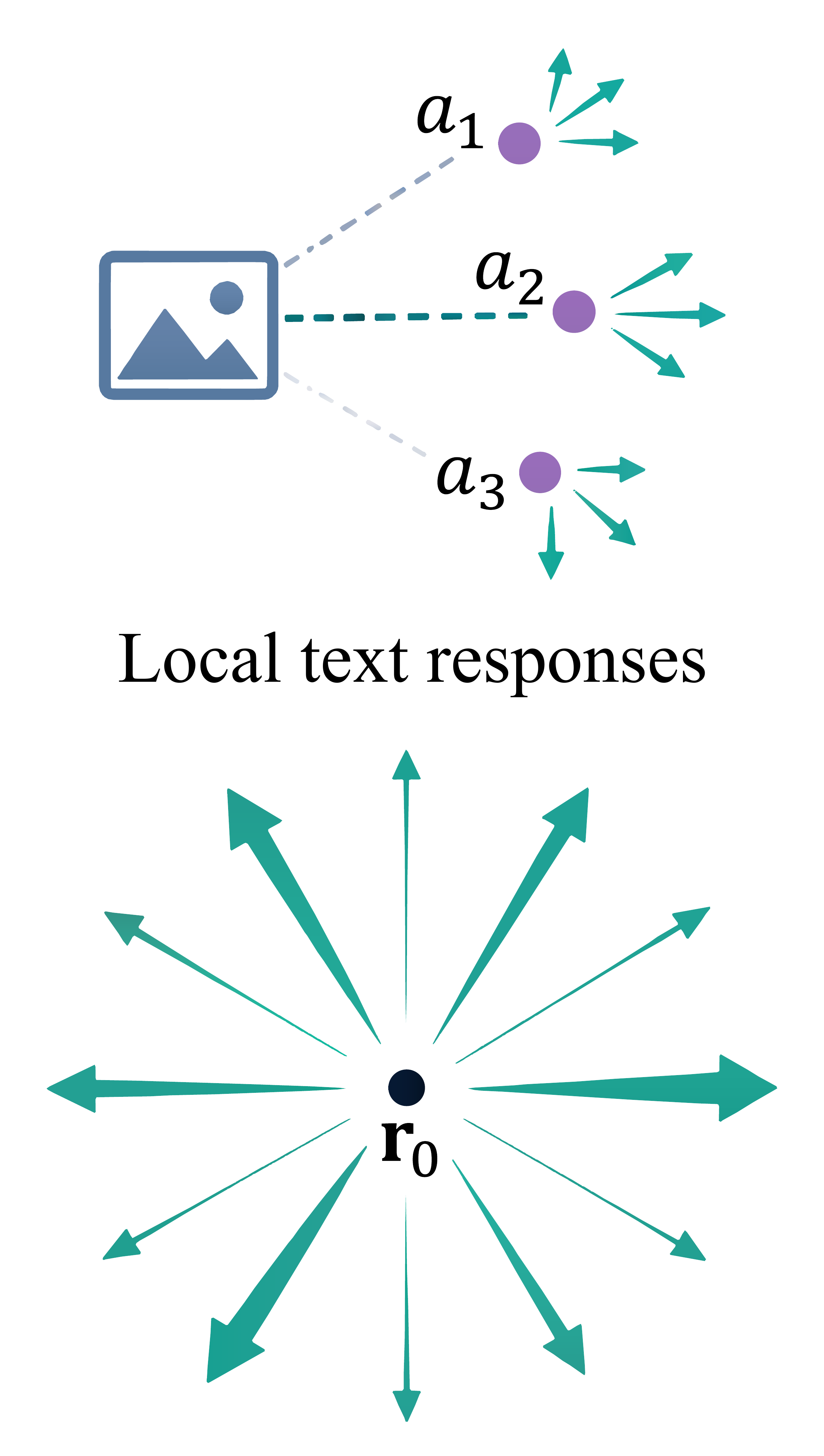}
    \caption{TRAG (ours)}
    \label{fig:motivation-c}
  \end{subfigure}
  \par\vspace{-4pt}
  \caption{\textbf{Directional constraints on visual representation drift.}
    Thicker lower arrows indicate stronger penalties.
    (a) Feature matching constrains drift equally in all directions.
    (b) Prediction distribution matching weights directions associated with differences between candidate textual representations using pre-trained prediction probabilities.
    (c) TRAG constructs anchor-level constraints from local responses of the pre-trained text encoder and aggregates them using pre-trained image--anchor relevance.}
  \label{fig:motivation}
  \vspace{-4pt}
\end{figure}

Existing methods often use guidance from the pre-trained model to preserve image--text alignment during adversarial fine-tuning.
Some match visual features between the fine-tuned and pre-trained models~\cite{FARE,QTAFT}, whereas others match prediction distributions over candidate texts~\cite{PMGAFT,AGFT}.
Both forms of guidance aim to preserve pre-trained image--text relationships, which depend on projections of visual representations along different directions in the shared embedding space.
For normalized representations, projections onto textual representations determine individual image--text similarities, while projections onto differences between textual representations determine relative similarities~\cite{CLIP}.
Changes in visual representations along different directions can affect these relationships differently.
How guidance constrains these changes can therefore influence the preservation of image--text relationships.

From this perspective, we analyze both forms of guidance as directional constraints on drift from pre-trained visual representations, revealing complementary limitations.
As illustrated in Figs.~\ref{fig:motivation}\subref{fig:motivation-a} and~\ref{fig:motivation}\subref{fig:motivation-b}, matching visual features constrains drift equally in all directions, providing broad coverage but no semantic emphasis.
Matching prediction distributions instead prioritizes semantically relevant directions associated with differences between candidate textual representations, but remains limited by the candidate texts and prediction probabilities.
Zero-shot generalization, however, requires preserving diverse image--text relationships across unseen tasks.
These limitations can leave such relationships insufficiently protected and degrade zero-shot performance.
We therefore seek directional references for potential semantic variations to construct guidance with broad coverage and semantic emphasis.

To address these limitations, we propose \textbf{T}extual \textbf{R}esponse-\textbf{A}ware \textbf{G}uidance (\textbf{TRAG}), which uses local responses of the pre-trained text encoder to constrain visual representation drift.
Using known texts as semantic anchors, we examine its responses to local variations in their input embeddings.
As shown in Fig.~\ref{fig:motivation}\subref{fig:motivation-c}, the response directions and magnitudes characterize local changes in textual representations, providing directional references for potential semantic variations beyond the given texts.
With a fixed total weight for directional penalties, we derive anchor-level constraints by minimizing the average worst-case error in preserving image--text relationships associated with these local changes.
Finally, we aggregate them using pre-trained image--anchor relevance on clean images, yielding an image-adaptive constraint for adversarial fine-tuning.

Extensive experiments demonstrate TRAG's state-of-the-art clean and robust performance.
We validate these advantages across multiple VLM architectures on zero-shot classification and image--text retrieval, covering both class-supervised and image--text fine-tuning settings under multiple threat models.
These benefits extend to image captioning and visual question answering in LVLMs.
Further analyses and ablations show that TRAG helps preserve pre-trained image--text relationships during robust adaptation.
Our main contributions are as follows:
\begin{itemize}
  \item We analyze the directional constraints that feature matching and prediction distribution matching impose on visual representation drift, revealing complementary limitations in preserving pre-trained image--text relationships.
  \item We propose TRAG, which derives anchor-level constraints from local responses of the pre-trained text encoder and aggregates them using image--anchor relevance, yielding an image-adaptive constraint.
  \item We demonstrate state-of-the-art clean and robust performance across architectures and tasks, with analyses and ablations showing that TRAG helps preserve pre-trained image--text relationships.
\end{itemize}

\section{Related Work}
\label{sec:related_work}
\vspace{-2pt}

\noindent\textbf{Adversarial robustness of VLMs.}
Adversarial attacks can induce incorrect predictions in deep neural networks through small, carefully crafted image perturbations, posing security risks~\cite{SzegedyAdversarialExamples,GoodfellowAdversarialExamples}.
Adversarial training is a widely adopted defense that uses min--max optimization to improve robustness, alternating between generating adversarial examples against the current model and updating its parameters~\cite{MadryAdversarialTraining}.
For VLMs, TeCoA~\cite{TeCoA} fine-tunes the image encoder with a text-guided contrastive objective to improve zero-shot adversarial robustness.
Subsequent methods increase separation between class textual representations~\cite{LAAT} or adjust classification margins~\cite{TIMA}.
Other approaches align regions along adversarial paths~\cite{AdvSimplex} or image--text subspaces~\cite{AoS}.
Adversarial prompt learning optimizes textual prompts~\cite{APT} or both textual and visual prompts~\cite{FAP} while keeping the pre-trained encoders fixed.
Test-time defenses improve inference robustness through prompt tuning~\cite{RTPT,TAPT} or input counterattacks~\cite{TTC}.

\noindent\textbf{Pre-trained model guidance.}
To mitigate degradation in zero-shot generalization, existing methods use pre-trained model guidance to preserve image--text alignment during adversarial fine-tuning.
At the feature level, FARE~\cite{FARE} matches adversarial visual features to the pre-trained model's clean features.
QT-AFT~\cite{QTAFT} combines feature matching with richer textual supervision from high-quality image captions.
At the prediction level, PMG-AFT~\cite{PMGAFT} combines an adversarial task objective with prediction distribution matching against the pre-trained model.
AGFT~\cite{AGFT} replaces hard-label supervision with the pre-trained model's soft alignment distributions, calibrated through temperature scaling.
TGA-ZSR~\cite{TGAZSR} preserves text-guided spatial attention using the original model as a reference.
AdvFLYP~\cite{AdvFLYP} combines contrastive adversarial fine-tuning on web image--text pairs with feature matching and prediction distribution matching.
We analyze the directional constraints imposed by representative feature and prediction distribution matching objectives.
TRAG uses local responses of the pre-trained text encoder to construct an image-adaptive constraint on visual representation drift.

\par\vspace{-3pt}
\section{Preliminaries and Analysis}
\label{sec:preliminaries_analysis}
\vspace{-2pt}

We introduce the basic setting and analyze how two representative forms of pre-trained guidance constrain visual representation drift.
Geometric diagnostics further examine their complementary limitations and inform guidance for preserving diverse image--text relationships.

\par\vspace{-3pt}
\subsection{Preliminaries}
\label{sec:preliminaries}
\vspace{-2pt}

\noindent\textbf{Vision-language models.}
VLMs such as CLIP~\cite{CLIP} use an image encoder $f_\theta:\mathcal X\to\mathbb R^d$ and a text encoder $g:\mathcal T\to\mathbb R^d$ to map images and texts into a shared embedding space, where $\mathcal X$ and $\mathcal T$ denote the image and text input spaces.
We update only the image encoder parameters $\theta$ during fine-tuning and use the original pre-trained image encoder $f_0$ as a frozen reference.
The pre-trained text encoder $g$ is shared by both models and remains fixed.
For an image $x$ and a text input $t$, their normalized representations are
\begin{equation}
\label{eq:normalized_representations}
\mathbf r_\theta(x)
=
\frac{f_\theta(x)}{\lVert f_\theta(x)\rVert_2},
\qquad
\mathbf s(t)
=
\frac{g(t)}{\lVert g(t)\rVert_2}.
\end{equation}
Their inner product gives image--text similarity.
For candidate texts $\mathcal T_{\mathrm{cand}}=\{t_j\}_{j=1}^{C}$, the prediction distribution is
\begin{equation}
\label{eq:candidate_prediction}
p_\theta(j\mid x;\mathcal T_{\mathrm{cand}})
=
\frac{
\exp\!\left(\mathbf r_\theta(x)^\top\mathbf s(t_j)/\tau\right)
}{
\sum_{k=1}^{C}
\exp\!\left(\mathbf r_\theta(x)^\top\mathbf s(t_k)/\tau\right)
},
\end{equation}
where $\tau>0$ is the temperature inherited from pre-training.
The pre-trained model's normalized visual representation $\mathbf r_0(x)$ and prediction distribution $p_0(j\mid x;\mathcal T_{\mathrm{cand}})$ are obtained by replacing $f_\theta$ with $f_0$.

\noindent\textbf{Adversarial fine-tuning.}
Given a clean image $x$, an adversarial attack seeks a perturbation $\delta$ that increases the model's loss, subject to $x^{\mathrm{adv}}=x+\delta\in B(x,\epsilon)$.
Here, $B(x,\epsilon)=\{x'\in\mathcal X:\lVert x'-x\rVert_\infty\le\epsilon\}$, where $\epsilon$ denotes the perturbation budget.
Adversarial fine-tuning updates the image encoder of a pre-trained VLM using these adversarial examples.
TeCoA~\cite{TeCoA} uses the following text-guided contrastive loss for a mini-batch $\{x_i\}$ from ImageNet~\cite{ImageNet}:
\begin{equation}
\label{eq:tecoa_loss}
\begin{aligned}[t]
&\mathcal L_{\mathrm{TeCoA}}
\bigl(\theta;\{x_i^{\mathrm{adv}}\},\mathcal T_{\mathrm{cand}},\{y_{ij}\}\bigr)
\\
&=-\mathbb E_{i,j}\left[
y_{ij}\log
\frac{
\exp\!\left(\mathbf r_\theta(x_i^{\mathrm{adv}})^\top\mathbf s(t_j)/\tau\right)
}{
\sum_{k=1}^{C}\exp\!\left(\mathbf r_\theta(x_i^{\mathrm{adv}})^\top\mathbf s(t_k)/\tau\right)
}
\right],
\end{aligned}
\end{equation}
where $y_{ij}=1$ indicates a positive image--text pair $(x_i,t_j)$, and $y_{ij}=0$ otherwise.
The min--max objective is
\begin{equation}
\label{eq:tecoa_minmax}
\min_\theta\mathbb E_{\{x_i\}\sim\mathcal D}
\Bigl[\max_{x_i^{\mathrm{adv}}\in B(x_i,\epsilon)}
\mathcal L_{\mathrm{TeCoA}}\Bigr],
\end{equation}
where $\mathcal D$ is the source dataset.
The inner maximization finds adversarial images; the outer minimization updates $\theta$.
The resulting model is evaluated zero-shot on downstream tasks without task-specific adaptation.

\begin{figure}[t]
  \centering
  \includegraphics[width=\linewidth]{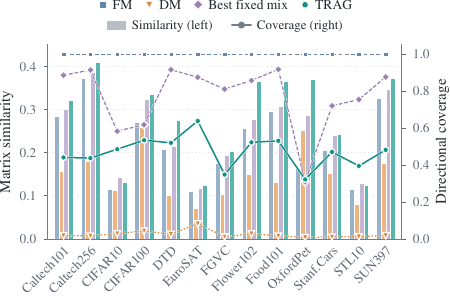}
  \captionsetup{skip=7pt}
  \caption{\textbf{Geometric diagnostics of \mbox{CLIP ViT-B/32} before fine-tuning.}
  Bars (left) show similarity to downstream class geometry; lines (right) show directional coverage.
  The best fixed FM--DM mix is selected by mean similarity per dataset.}
  \label{fig:geometric_diagnostics}
  \vspace{-3pt}
\end{figure}

\subsection{Directional Analysis of Guidance}
\label{sec:directional_analysis}

To understand how pre-trained guidance preserves image--text relationships, we analyze two representative forms: feature matching (FM) and prediction distribution matching (DM).
Both constrain visual representation drift from the original model's clean representation:
\begin{equation}
\label{eq:visual_representation_drift}
\Delta\mathbf r(x,x^{\mathrm{adv}})
=
\mathbf r_\theta(x^{\mathrm{adv}})-\mathbf r_0(x).
\end{equation}
This drift reflects both model adaptation and input perturbation.
It changes individual image--text similarities by $\mathbf s(t)^\top\Delta\mathbf r$ and relative similarities to $t_i$ versus $t_j$ by $[\mathbf s(t_i)-\mathbf s(t_j)]^\top\Delta\mathbf r$.
Drift along different directions can therefore affect image--text relationships differently.

\noindent\textbf{Feature matching.}
FM encourages current visual representations to remain close to those of the pre-trained model~\cite{FARE,QTAFT}.
We analyze the normalized form:
\begin{equation}
\label{eq:feature_matching}
\mathcal L_{\mathrm{FM}}
=
\lVert\mathbf r_\theta(x^{\mathrm{adv}})-\mathbf r_0(x)\rVert_2^2
=
\Delta\mathbf r^\top\mathbf I_d\Delta\mathbf r.
\end{equation}
The identity matrix constrains drift equally in all directions.
This provides broad protection for image--text relationships without favoring semantically relevant directions.

\noindent\textbf{Prediction distribution matching.}
DM uses pre-trained predictions as soft supervision~\cite{PMGAFT,AGFT}.
We analyze forward KL $D_{\mathrm{KL}}(\mathbf p_0\|\mathbf p_\theta)$ over fixed candidate texts $\mathcal T_{\mathrm{cand}}$, where $\mathbf p_0$ and $\mathbf p_\theta$ denote the original clean and current adversarial distributions, respectively.
Around $\mathbf r_0(x)$,
\begin{equation}
\label{eq:distribution_matching_local}
\mathcal L_{\mathrm{DM}}
=
\frac12\Delta\mathbf r^\top\mathbf H_0\Delta\mathbf r
+
O(\lVert\Delta\mathbf r\rVert_2^3),
\end{equation}
where $\mathbf s_i=\mathbf s(t_i)$ and the reference Hessian is
\begin{equation}
\label{eq:distribution_matching_hessian}
\mathbf H_0
=
\frac{1}{2\tau^2}
\sum_{i,j=1}^{C}
[\mathbf p_0]_i[\mathbf p_0]_j
(\mathbf s_i-\mathbf s_j)
(\mathbf s_i-\mathbf s_j)^\top.
\end{equation}
The outer products penalize drift along differences between candidate textual representations, with each pair weighted by $[\mathbf p_0]_i[\mathbf p_0]_j$.
DM thus provides semantic emphasis, but remains limited by the candidate texts and prediction probabilities.
Further derivations and connections to specific methods are provided in the appendix.

\begin{figure*}[t]
  \centering
  \includegraphics[width=\textwidth]{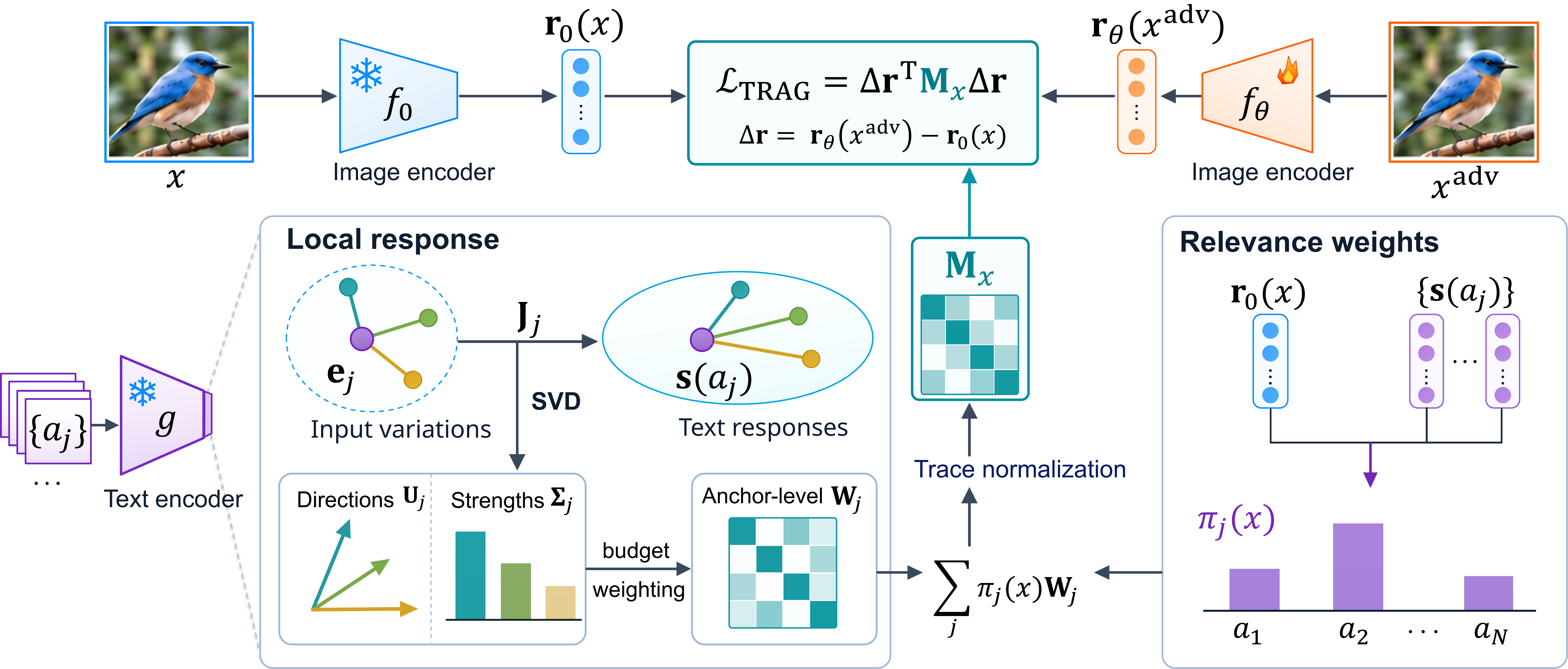}
  \caption{\textbf{Overview of TRAG.}
    We construct anchor-level constraints $\mathbf W_j$ from the text encoder's local response directions and magnitudes under a fixed-budget criterion.
    Pre-trained image--anchor relevance on clean images weights their aggregation.
    After global trace normalization, $\mathbf M_x$ constrains visual representation drift during adversarial fine-tuning.}
  \label{fig:trag_framework}
\end{figure*}

FM and DM thus exhibit complementary limitations.
Preserving the diverse image--text relationships needed for zero-shot transfer requires controlling drift along semantic directions relevant to unseen tasks.
These limitations can hinder such preservation across potential semantic variations and degrade zero-shot performance.
\par

\subsection{Geometric Diagnostics}
\label{sec:geometric_diagnostics}

To complement the directional analysis, we examine the matrices defining these constraints before fine-tuning.
We use the normalized participation ratio of their eigenvalues~\cite{LitwinKumar2017} to measure how broadly directional weights are distributed.
The lines in Fig.~\ref{fig:geometric_diagnostics} show maximal coverage for FM but low coverage for DM.
The eigenvalue spectra in Fig.~\ref{fig:constraint_spectra} illustrate DM's concentrated directional weights.

To assess relevance to downstream tasks, we use the covariance of normalized downstream class textual representations as a reference for task-relevant semantic directions.
We compare each constraint matrix with this covariance using the normalized Frobenius inner product.
The bars in Fig.~\ref{fig:geometric_diagnostics} show lower similarity for FM and DM than for TRAG.
Their best fixed mixtures generally have lower similarity despite often having greater coverage than TRAG.

Together, these analyses motivate guidance with broad coverage and semantic emphasis for preserving diverse image--text relationships across unseen tasks.
We therefore seek directional references for potential semantic variations to design constraints on visual representation drift.
\par

\section{Methodology}
\label{sec:methodology}

We describe the construction of TRAG and its use in adversarial fine-tuning, as illustrated in Fig.~\ref{fig:trag_framework}.
We first derive anchor-level constraints from local responses of the pre-trained text encoder under a fixed penalty budget (Sec.~\ref{sec:anchor_constraints}).
We then aggregate them using pre-trained image--anchor relevance and formulate the training objective (Sec.~\ref{sec:image_adaptive_guidance}).

\begin{table*}[t]
\centering
\caption{\textbf{Zero-shot classification with CLIP ViT-B/32.} Models are fine-tuned on ImageNet with training/test budgets of $2/255$. Bold and underlined values mark the best and second-best fine-tuned results; Average spans all 16 datasets.}
\label{tab:clean_robust}
\vspace{-9pt}
\resizebox{\linewidth}{!}{\setlength{\tabcolsep}{5pt}
\renewcommand{\arraystretch}{0.9}
\setlength{\heavyrulewidth}{1.0pt}
\setlength{\lightrulewidth}{0.7pt}
\begin{tabular}{cc*{16}{c}>{\columncolor[RGB]{220,236,250}}c}
\toprule
\textbf{Eval.} & \textbf{Method}
& \rotatebox{90}{Caltech101}
& \rotatebox{90}{Caltech256}
& \rotatebox{90}{CIFAR10}
& \rotatebox{90}{CIFAR100}
& \rotatebox{90}{DTD}
& \rotatebox{90}{EuroSAT}
& \rotatebox{90}{FGVC}
& \rotatebox{90}{Flower102}
& \rotatebox{90}{Food101}
& \rotatebox{90}{ImageNet}
& \rotatebox{90}{ImageNet-R}
& \rotatebox{90}{ImageNet-S}
& \rotatebox{90}{OxfordPet}
& \rotatebox{90}{Stanf.Cars}
& \rotatebox{90}{STL10}
& \rotatebox{90}{SUN397}
& \textbf{Average} \\
\midrule

& CLIP
& 88.71 & 84.66 & 87.55 & 62.56 & 41.38
& 43.72 & 19.17 & 66.15 & 83.51 & 61.89
& 66.26 & 40.75 & 87.33 & 58.62 & 97.28
& 60.10 & 65.60 \\

& TeCoA
& 82.89 & 72.92 & 73.45 & 40.84 & 30.74
& 19.59 & 11.25 & 42.86 & 41.83 & 55.66
& 49.28 & 27.54 & 75.23 & 25.28 & 90.22
& 47.67 & 49.20 \\

& PMG-AFT
& 86.46 & 77.32 & 80.13 & 50.87 & 31.65
& \textbf{22.90} & 12.51 & 45.22 & 52.50 & 61.19
& 54.91 & 33.74 & 79.54 & 32.75 & 93.34
& 50.24 & 54.08 \\

& FARE
& \underline{86.51} & \underline{79.72} & 78.40 & 57.50 & 34.04
& 17.59 & \textbf{14.94} & \textbf{52.23} & \textbf{64.72} & 56.87
& 57.53 & 35.61 & \underline{83.19} & \textbf{48.71} & 93.77
& \underline{52.05} & \underline{57.09} \\

& TGA-ZSR
& 82.84 & 76.13 & 80.75 & 49.01 & 28.72
& 14.10 & 9.78 & 43.92 & 45.16 & \underline{70.53}
& 57.33 & \underline{38.16} & 77.82 & 25.74 & 93.19
& 47.88 & 52.57 \\

& QT-AFT
& 86.18 & 78.54 & 82.47 & 56.50 & \underline{34.20}
& 11.13 & \underline{14.31} & \underline{51.32} & 62.08 & 55.12
& 56.13 & 34.47 & 81.91 & \underline{47.07} & 94.38
& 51.38 & 56.07 \\

& AGFT
& 86.22 & 78.83 & \underline{83.10} & \underline{58.22} & 33.24
& 19.29 & 13.47 & 45.59 & 59.22 & 59.85
& \underline{58.36} & 36.60 & 83.00 & 41.01 & \underline{95.25}
& 50.38 & 56.35 \\

\rowcolor[RGB]{220,236,250}
\cellcolor{white}\multirow{-8}{*}{\rotatebox{90}{\textbf{Clean}}}
& \textbf{TRAG (Ours)}
& \textbf{87.49} & \textbf{81.20} & \textbf{86.01}
& \textbf{62.32} & \textbf{35.00}
& \underline{22.80} & 13.47 & 49.93 & \underline{62.78}
& \textbf{71.16} & \textbf{61.49} & \textbf{41.37}
& \textbf{84.20} & 45.29 & \textbf{95.69}
& \textbf{53.39} & \textbf{59.60} \\

\midrule

& CLIP
& 0.61 & 0.18 & 0.11 & 0.57 & 0.37
& 0.46 & 0.63 & 0.55 & 0.11 & 0.27
& 0.45 & 0.17 & 0.08 & 0.50 & 0.12
& 0.45 & 0.35 \\

& TeCoA
& 60.23 & 46.48 & 42.25 & 20.88 & 17.23
& 10.93 & 2.10 & 16.78 & 11.41 & 23.25
& 24.26 & 14.31 & 38.15 & 6.06 & 69.15
& 17.74 & 26.33 \\

& PMG-AFT
& 66.43 & 54.00 & 51.21 & 27.62 & 20.00
& \textbf{11.87} & 3.21 & 18.11 & 17.73 & 32.13
& 30.66 & 18.72 & 52.13 & 9.34 & 76.75
& \underline{21.93} & 31.99 \\

& FARE
& 61.81 & 51.49 & 49.13 & 27.71 & 19.36
& 7.67 & 2.94 & 16.46 & 17.74 & 23.80
& 27.94 & 16.64 & 44.74 & 9.04 & 74.44
& 18.53 & 29.34 \\

& TGA-ZSR
& 63.14 & 50.21 & 49.10 & 24.27 & 14.84
& 0.78 & 1.95 & 15.54 & 12.71 & \underline{38.72}
& 29.62 & 20.34 & 49.48 & 5.73 & 74.59
& 18.09 & 29.32 \\

& QT-AFT
& 62.86 & 51.94 & 52.69 & 29.04 & 20.37
& 1.60 & 2.58 & 17.35 & 18.46 & 24.41
& 28.59 & 16.85 & 45.56 & 9.16 & 77.14
& 19.17 & 29.86 \\

& AGFT
& \underline{66.77} & \underline{55.82} & \underline{53.56}
& \underline{30.87} & \underline{20.48}
& 11.04 & \textbf{4.83} & \underline{18.52}
& \underline{19.86} & 32.77
& \underline{32.97} & \underline{20.54}
& \underline{55.64} & \textbf{12.16}
& \underline{79.49} & \textbf{22.35}
& \underline{33.60} \\

\rowcolor[RGB]{220,236,250}
\cellcolor{white}\multirow{-8}{*}{\rotatebox{90}{\textbf{Robust}}}
& \textbf{TRAG (Ours)}
& \textbf{67.11} & \textbf{57.37} & \textbf{56.82}
& \textbf{33.69} & \textbf{21.44}
& \underline{11.24} & \underline{3.45}
& \textbf{19.01} & \textbf{19.87}
& \textbf{41.48} & \textbf{34.25} & \textbf{22.68}
& \textbf{58.86} & \underline{11.40}
& \textbf{80.99} & 21.84 & \textbf{35.09} \\

\bottomrule
\end{tabular}}
\end{table*}

\subsection{Anchor-Level Constraint Construction}
\label{sec:anchor_constraints}

Zero-shot transfer requires preserving diverse image--text relationships for downstream tasks unknown during fine-tuning.
Known texts provide semantic reference points but may not fully capture these needs.
Starting from known concepts, we therefore seek directional references for potential semantic variations.
The pre-trained text encoder's learned mapping translates local input embedding changes into textual representation changes in the shared space.
Using these texts as semantic anchors, we construct drift constraints from local response directions and magnitudes.

\noindent\textbf{Local textual responses.}
Let $\mathcal A=\{a_j\}_{j=1}^{N}$ denote the anchor set.
For anchor $a_j$, $\mathbf e_j\in\mathbb R^{n_jd_e}$ concatenates the embeddings of its $n_j$ body tokens, excluding start and end tokens, where $d_e$ is the token embedding dimension.
With the encoder frozen and other inputs fixed, its normalized-output Jacobian has a thin singular value decomposition:
\begin{equation}
\label{eq:anchor_jacobian_svd}
\mathbf J_j
=\frac{\partial\mathbf s(a_j)}{\partial\mathbf e_j}
=\mathbf U_j\boldsymbol\Sigma_j\mathbf V_j^\top.
\end{equation}
Left singular vectors describe response directions in the shared space; singular values quantify their magnitudes.

For a small embedding change $\Delta\mathbf e_j$, the linear response $\Delta\mathbf s_j=\mathbf J_j\Delta\mathbf e_j$ approximates the textual representation difference from the anchor.
Following Sec.~\ref{sec:directional_analysis}, visual drift $\Delta\mathbf r$ changes relative similarity to the locally varied representation versus the anchor by approximately $\Delta\mathbf s_j^\top\Delta\mathbf r$.
We use $\Delta\mathbf s_j$ as a local proxy for potential semantic variations.

\noindent\textbf{Directional penalty allocation.}
We model $\Delta\mathbf e_j$ as a random input proxy.
The response second moment $\mathbf K_j=\mathbb E[\Delta\mathbf s_j\Delta\mathbf s_j^\top]$ summarizes average squared responses across directions.
We choose a positive semidefinite matrix $\mathbf Q$ to penalize drift through $\Delta\mathbf r^\top\mathbf Q\Delta\mathbf r$.
Its eigenvalues are directional penalty weights, with a fixed sum $\operatorname{tr}(\mathbf Q)=d$.

For each response, $\Delta\mathbf s_j^\top\Delta\mathbf r$ measures the error in relative image--text similarity under the linear proxy model.
Using a common unit penalty bound, we take the largest squared error over abstract drifts and then average over responses:
\begin{equation}
\label{eq:proxy_relation_risk}
\mathcal R(\mathbf Q;\mathbf K_j)
=\mathbb E_{\Delta\mathbf s_j}
\left[
\max_{\Delta\mathbf r^\top\mathbf Q\Delta\mathbf r\leq 1}
(\Delta\mathbf s_j^\top\Delta\mathbf r)^2
\right].
\end{equation}

\newcounter{proposition}
\begin{samepage}
\noindent\refstepcounter{proposition}\label{prop:budget_allocation}\textbf{Proposition~\theproposition\ (Budget allocation).}
Suppose $\Delta\mathbf s_j$ has a finite, nonzero second moment.
For $\mathbf Q\succeq0$ satisfying $\operatorname{range}(\mathbf K_j)\subseteq\operatorname{range}(\mathbf Q)$, the risk is $\mathcal R(\mathbf Q;\mathbf K_j)=\operatorname{tr}(\mathbf K_j\mathbf Q^\dagger)$, where $\mathbf Q^\dagger$ denotes the Moore--Penrose pseudoinverse~\cite{Penrose1955}.
Under $\operatorname{tr}(\mathbf Q)=d$, a minimizer is
\begin{equation}
\label{eq:single_anchor_budget_optimum}
\mathbf Q^\star
=\frac{d\sqrt{\mathbf K_j}}{\operatorname{tr}(\sqrt{\mathbf K_j})}.
\end{equation}
The proof is provided in the appendix.
\end{samepage}

We use a zero-mean isotropic proxy with $\operatorname{Cov}(\Delta\mathbf e_j)=\mathbf I_{n_jd_e}/n_j$.
Equal variance across input directions lets the pre-trained text encoder determine the response geometry, while $1/n_j$ keeps the expected total input variation energy independent of the body-token count.
This gives
\begin{equation}
\label{eq:anchor_response_second_moment}
\mathbf K_j
=\mathbb E[\Delta\mathbf s_j\Delta\mathbf s_j^\top]
=\frac{\mathbf J_j\mathbf J_j^\top}{n_j}.
\end{equation}
Proposition~\ref{prop:budget_allocation} yields the anchor-level constraint shape
\begin{equation}
\label{eq:anchor_constraint_matrix}
\mathbf W_j
=\sqrt{\mathbf K_j}
=\frac{\mathbf U_j\boldsymbol\Sigma_j\mathbf U_j^\top}{\sqrt{n_j}}.
\end{equation}
The penalty $\Delta\mathbf r^\top\mathbf W_j\Delta\mathbf r$ weights squared drift projections onto response directions by singular values scaled by $1/\sqrt{n_j}$.
The single-anchor optimum $\mathbf Q^\star$ rescales $\mathbf W_j$ to trace $d$.
We retain the unnormalized $\mathbf W_j$ to preserve response strengths across anchors; Sec.~\ref{sec:image_adaptive_guidance} allocates their scales jointly under a shared budget before aggregation.

\subsection{Image-Adaptive Guidance and Fine-Tuning}
\label{sec:image_adaptive_guidance}

\noindent\textbf{Image-adaptive aggregation.}
Although downstream tasks are unknown, a clean image $x$ provides cues about which concept neighborhoods are relevant.
We therefore use pre-trained image--anchor relevance as a proxy for their relative importance:
\begin{equation}
\label{eq:anchor_relevance}
\pi_j(x)
=\frac{\exp\!\left(\mathbf r_0(x)^\top\mathbf s(a_j)/\tau\right)}
{\sum_{k=1}^{N}\exp\!\left(\mathbf r_0(x)^\top\mathbf s(a_k)/\tau\right)}.
\end{equation}

For fixed $x$, let $\mathbf Q_j$ denote component constraint matrices satisfying the semidefiniteness and range conditions of Proposition~\ref{prop:budget_allocation}.
We minimize the weighted proxy risk $\sum_{j=1}^{N}\pi_j(x)\mathcal R(\mathbf Q_j;\mathbf K_j)$, subject to the shared budget $\sum_{j=1}^{N}\pi_j(x)\operatorname{tr}(\mathbf Q_j)=d$.
A closed-form solution scales all $\mathbf W_j$ by a common factor, as derived in the appendix.
Aggregating these components with weights $\pi_j(x)$ gives:
\begin{equation}
\label{eq:image_adaptive_matrix}
\mathbf M_x
=\frac{d\sum_{j=1}^{N}\pi_j(x)\mathbf W_j}
{\sum_{j=1}^{N}\pi_j(x)\operatorname{tr}(\mathbf W_j)}.
\end{equation}
The common normalization fixes $\operatorname{tr}(\mathbf M_x)=d$ while preserving relative response strengths across anchors.

\noindent\textbf{Adversarial fine-tuning.}
For a mini-batch $\{(x_i,t_i)\}$, where $t_i$ denotes a paired text or the corresponding class text, write $\Delta\mathbf r_i=\Delta\mathbf r(x_i,x_i^{\mathrm{adv}})$.
We average the TRAG penalty over images:
\begin{equation}
\label{eq:trag_guidance_loss}
\mathcal L_{\mathrm{TRAG}}
=\mathbb E_i
\left[
\Delta\mathbf r_i^\top\mathbf M_{x_i}\Delta\mathbf r_i
\right].
\end{equation}
The task loss $\mathcal L_{\mathrm{task}}$ can be $\mathcal L_{\mathrm{TeCoA}}$ for classification or InfoNCE~\cite{InfoNCE} for paired image--text training.
We optimize
\begin{equation}
\label{eq:trag_minmax}
\min_\theta
\mathbb E_{\{(x_i,t_i)\}\sim\mathcal D}
\left[
\max_{x_i^{\mathrm{adv}}\in B(x_i,\epsilon)}
\left(
\mathcal L_{\mathrm{task}}
+\lambda\mathcal L_{\mathrm{TRAG}}
\right)
\right],
\end{equation}
where $\lambda\geq0$ controls the relative guidance strength.

We approximate the inner maximization over the batch images using projected gradient descent (PGD)~\cite{MadryAdversarialTraining}.
Both the attack and outer update use the same combined loss.
Only the image encoder is updated; the original model, text encoder, and temperature remain frozen.
The clean references $\mathbf r_0(x_i)$, relevance weights $\pi_j(x_i)$, and matrices $\mathbf M_{x_i}$ remain fixed throughout each attack.

We precompute anchor representations $\mathbf s(a_j)$ and matrices $\mathbf W_j$ once before fine-tuning.
Inference uses the adapted image encoder within the original VLM pipeline, with no additional computation from TRAG.

\section{Experiments}
\label{sec:experiments}

\begin{table}[t]
\caption{\textbf{Different test budgets.} ViT-B/32 models are trained at $2/255$. Results are means over 16 datasets; Avg. further averages robust accuracy over all four listed test budgets.}
\vspace{-9pt}
\centering
\footnotesize
\setlength{\tabcolsep}{2pt}
\renewcommand{\arraystretch}{0.9}

\begin{tabularx}{0.927\columnwidth}{
    c*{6}{>{\centering\arraybackslash}X}
}
\toprule
\multirow{2}{*}[-0.6ex]{Method}
& \multirow{2}{*}[-0.6ex]{Clean}
& \multicolumn{5}{c}{Robust Accuracy} \\
\cmidrule(lr){3-7}
& & 1/255 & 2/255 & 3/255 & 4/255 & Avg. \\
\midrule

TeCoA
& 49.20 & 37.42 & 26.33 & 17.40 & 10.75 & 22.97 \\

PMG-AFT
& 54.08 & 42.88 & 31.99 & 22.26 & 14.53 & 27.92 \\

FARE
& \underline{57.09} & 42.81 & 29.34 & 18.33 & 10.76 & 25.31 \\

TGA-ZSR
& 52.57 & 40.50 & 29.32 & 19.47 & 11.69 & 25.24 \\

QT-AFT
& 56.07 & 42.52 & 29.86 & 19.33 & 11.77 & 25.87 \\

AGFT
& 56.35 & \underline{44.83} & \underline{33.60}
& \underline{23.44} & \underline{15.13}
& \underline{29.25} \\

\rowcolor[RGB]{220,236,250}
\textbf{TRAG (Ours)}
& \textbf{59.60} & \textbf{47.07} & \textbf{35.09}
& \textbf{24.25} & \textbf{15.38}
& \textbf{30.45} \\

\bottomrule
\end{tabularx}
\label{tab:eps_compare}
\end{table}

\begin{table}[t]
\caption{\textbf{Different training budgets.} Mean accuracy (\%) over 16 datasets; ViT-B/32 uses matched training/test budgets.}
\vspace{-9pt}
\centering
\footnotesize
\setlength{\tabcolsep}{2pt}
\renewcommand{\arraystretch}{0.9}

\begin{tabularx}{0.927\columnwidth}{
    c*{6}{>{\centering\arraybackslash}X}
}
\toprule
\multirow{2}{*}[-0.6ex]{Method}
& \multicolumn{2}{c}{1/255}
& \multicolumn{2}{c}{3/255}
& \multicolumn{2}{c}{4/255} \\
\cmidrule(lr){2-3}
\cmidrule(lr){4-5}
\cmidrule(lr){6-7}
& Clean & Robust & Clean & Robust & Clean & Robust \\
\midrule

TeCoA
& 56.31 & 38.43 & 42.87 & 18.87 & 37.45 & 13.95 \\

PMG-AFT
& 59.58 & 42.72 & 46.29 & 22.63 & 42.01 & 17.59 \\

FARE
& \underline{61.90} & 38.66 & 52.46 & 22.80
& 47.74 & 17.88 \\

TGA-ZSR
& 58.37 & 37.66 & 47.72 & 23.16
& 43.86 & 18.71 \\

QT-AFT
& 61.03 & 39.55 & 52.06 & 23.40
& 48.00 & 18.66 \\

AGFT
& 60.94 & \underline{43.18}
& \underline{52.54} & \underline{26.55}
& \underline{48.97} & \underline{21.41} \\

\rowcolor[RGB]{220,236,250}
\textbf{TRAG (Ours)}
& \textbf{63.66} & \textbf{44.39}
& \textbf{56.35} & \textbf{28.12}
& \textbf{54.55} & \textbf{22.18} \\

\bottomrule
\end{tabularx}
\label{tab:eps_generalization}
\end{table}

\begin{table}[t]
\caption{\textbf{Different backbones.} Mean accuracy (\%) over 16 datasets; training and test budgets are $2/255$.}
\vspace{-9pt}
\centering
\footnotesize
\setlength{\tabcolsep}{2pt}
\renewcommand{\arraystretch}{0.9}

\begin{tabularx}{0.927\columnwidth}{
    c*{6}{>{\centering\arraybackslash}X}
}
\toprule
\multirow{2}{*}[-0.6ex]{Method}
& \multicolumn{2}{c}{ViT-B/16}
& \multicolumn{2}{c}{ViT-L/14}
& \multicolumn{2}{c}{RN101} \\
\cmidrule(lr){2-3}
\cmidrule(lr){4-5}
\cmidrule(lr){6-7}
& Clean & Robust & Clean & Robust & Clean & Robust \\
\midrule

TeCoA
& 53.38 & 30.18 & 64.12 & 39.99 & 31.65 & 16.45 \\

PMG-AFT
& 55.40 & 30.83 & 63.85 & 43.40 & 47.01 & 27.69 \\

FARE
& \underline{60.28} & 32.69
& \underline{69.06} & 45.03
& \underline{51.87} & 20.26 \\

TGA-ZSR
& 51.66 & 32.16 & 60.82 & 40.34 & -- & -- \\

QT-AFT
& 60.05 & 33.35 & 69.05 & 45.36
& 45.46 & 21.64 \\

AGFT
& 58.68 & \underline{36.25}
& 67.37 & \underline{46.78}
& 51.15 & \underline{28.82} \\

\rowcolor[RGB]{220,236,250}
\textbf{TRAG (Ours)}
& \textbf{62.95} & \textbf{39.08}
& \textbf{70.78} & \textbf{49.54}
& \textbf{55.39} & \textbf{29.54} \\

\bottomrule
\end{tabularx}
\label{tab:backbone_comparison}
\end{table}

\subsection{Zero-Shot Image Classification}
\label{sec:classification}

\noindent\textbf{Experimental setup.}
We fine-tune CLIP ViT-B/32 on ImageNet~\cite{ImageNet} and evaluate on ImageNet and 15 zero-shot datasets: Caltech-101~\cite{Caltech101}, Caltech-256~\cite{Caltech256}, CIFAR-10~\cite{CIFAR}, CIFAR-100~\cite{CIFAR}, DTD~\cite{DTD}, EuroSAT~\cite{EuroSAT}, FGVC Aircraft~\cite{FGVCAircraft}, Flowers102~\cite{Flowers102}, Food-101~\cite{Food101}, ImageNet-R~\cite{ImageNetR}, ImageNet-S~\cite{ImageNetSketch}, Oxford-IIIT Pet~\cite{OxfordIIITPet}, Stanford Cars~\cite{StanfordCars}, STL-10~\cite{STL10}, and SUN397~\cite{SUN397}.
Unless otherwise stated, training and evaluation use an $\ell_\infty$ budget of $2/255$.
For TRAG, we use five epochs, 5-step PGD with step size $1/255$, and $\lambda=15$.
The 1,000 ImageNet class names serve as anchors.

We reproduce TeCoA~\cite{TeCoA}, PMG-AFT~\cite{PMGAFT}, FARE~\cite{FARE}, TGA-ZSR~\cite{TGAZSR}, QT-AFT~\cite{QTAFT}, and AGFT~\cite{AGFT} using their respective training configurations.
All methods share the evaluation protocol.
Following FARE in attack selection, we use APGD-CE and targeted APGD-DLR from AutoAttack~\cite{AutoAttack}, with 20 iterations each.
We report per-dataset accuracy and averages across the 16 datasets.
Further details are provided in the appendix.

\noindent\textbf{Main results.}
Table~\ref{tab:clean_robust} shows that TRAG achieves the highest average clean and robust accuracy among fine-tuned methods: 59.60\% and 35.09\%, respectively.
These exceed the strongest baseline for each metric by 2.51 and 1.49 percentage points.
TRAG leads on 11 datasets for clean accuracy and 12 for robust accuracy, covering both ImageNet and zero-shot benchmarks.
For instance, TRAG improves CIFAR-100 clean and robust classification accuracy over AGFT by 4.10 and 2.82 percentage points, respectively.

\noindent\textbf{Different test budgets.}
Table~\ref{tab:eps_compare} evaluates checkpoints trained at $2/255$ under test budgets from $1/255$ to $4/255$.
TRAG leads in mean robust accuracy at all four budgets, including those above its training budget.
The margin over AGFT narrows from 2.24 points at $1/255$ to 0.25 points at $4/255$, showing smaller margins at larger budgets.

\noindent\textbf{Different training budgets.}
We also train separate models at $1/255$, $3/255$, and $4/255$ and evaluate each at its training budget (Table~\ref{tab:eps_generalization}).
TRAG attains the best mean clean and robust accuracy in all three settings with $\lambda$ fixed.
At $4/255$, it outperforms the strongest baseline, AGFT, by 5.58 points in clean accuracy and 0.77 points in robust accuracy.

\begin{table*}[t]
\centering
\caption{\textbf{Zero-shot image--text retrieval after CC3M fine-tuning.} TR/IR denote image-to-text/text-to-image recall (\%). Robust results use image PGD at $2/255$. Bold/underlined values mark the best/second-best fine-tuned results.}
\label{tab:retrieval_clean_robust}
\vspace{-9pt}
\resizebox{\linewidth}{!}{\setlength{\tabcolsep}{3pt}
\renewcommand{\arraystretch}{0.9}
\setlength{\heavyrulewidth}{1.0pt}
\setlength{\lightrulewidth}{0.7pt}

\begin{tabular}{cc*{16}{c}}
\toprule
\multirow{3}{*}{\textbf{Model}}
& \multirow{3}{*}{\textbf{Method}}
& \multicolumn{8}{c}{\textbf{Flickr30k}}
& \multicolumn{8}{c}{\textbf{COCO}} \\
\cmidrule(lr){3-10}
\cmidrule(lr){11-18}

& & \multicolumn{4}{c}{\textbf{Clean}}
& \multicolumn{4}{c}{\textbf{Robust}}
& \multicolumn{4}{c}{\textbf{Clean}}
& \multicolumn{4}{c}{\textbf{Robust}} \\
\cmidrule(lr){3-6}
\cmidrule(lr){7-10}
\cmidrule(lr){11-14}
\cmidrule(lr){15-18}

& & TR@1 & TR@10 & IR@1 & IR@10
& TR@1 & TR@10 & IR@1 & IR@10
& TR@1 & TR@10 & IR@1 & IR@10
& TR@1 & TR@10 & IR@1 & IR@10 \\
\midrule

& CLIP
& 82.60 & 99.10 & 62.08 & 92.02
& 0.30 & 1.40 & 0.16 & 1.08
& 53.18 & 85.14 & 33.32 & 68.95
& 0.10 & 0.36 & 0.04 & 0.30 \\

& TeCoA
& 69.20 & 92.90 & 55.96 & 87.44
& 45.00 & 75.40 & 34.44 & 71.26
& 41.32 & 76.00 & 28.02 & 64.09
& 21.90 & 52.80 & 15.09 & 43.73 \\

& PMG-AFT
& 76.00 & 95.50 & 59.44 & 89.16
& 54.70 & 83.20 & 42.20 & 77.08
& 45.58 & 79.24 & 30.68 & 66.93
& 29.72 & 62.22 & 19.10 & 51.55 \\

& FARE
& 78.70 & 96.90 & 61.62 & 90.42
& 52.50 & 84.80 & 39.34 & 76.24
& 45.04 & 80.62 & 31.28 & 67.16
& 25.66 & 58.90 & 17.41 & 48.40 \\

& TGA-ZSR
& 75.10 & 96.10 & 61.74 & 91.08
& 49.30 & 80.70 & 38.58 & 75.90
& 45.92 & 79.94 & 33.16 & 69.68
& 25.56 & 57.18 & 17.16 & 48.98 \\

& QT-AFT
& \underline{80.60} & \underline{97.00}
& \underline{64.48} & \underline{91.66}
& 55.10 & 85.20
& \underline{44.80} & \underline{79.26}
& \underline{46.10} & \underline{81.26}
& \underline{34.27} & \underline{70.50}
& 28.12 & 61.70
& \underline{20.15} & \underline{52.94} \\

& AGFT
& 77.20 & 96.20 & 60.70 & 89.38
& \underline{56.80} & \underline{86.60}
& 43.02 & 78.50
& 45.84 & 80.00 & 31.31 & 67.50
& \underline{29.78} & \underline{63.78}
& 19.22 & 52.57 \\

& AdvFLYP
& 75.90 & 95.90 & 60.38 & 89.74
& 54.00 & 84.70 & 42.90 & 77.64
& 44.64 & 79.06 & 31.52 & 67.90
& 27.38 & 60.72 & 19.22 & 51.72 \\

\rowcolor[RGB]{220,236,250}
\cellcolor{white}
\multirow{-9}{*}{\rotatebox{90}{\textbf{CLIP ViT-B/16}}}
& \textbf{TRAG (Ours)}
& \textbf{81.10} & \textbf{97.20}
& \textbf{65.78} & \textbf{92.36}
& \textbf{58.70} & \textbf{88.70}
& \textbf{46.72} & \textbf{81.34}
& \textbf{49.20} & \textbf{83.10}
& \textbf{35.77} & \textbf{71.70}
& \textbf{30.96} & \textbf{64.92}
& \textbf{21.63} & \textbf{55.34} \\

\midrule

& CoCa
& 89.10 & 99.90 & 74.52 & 95.52
& 6.90 & 19.70 & 3.94 & 17.28
& 62.88 & 89.96 & 45.60 & 79.06
& 3.34 & 10.90 & 1.86 & 9.19 \\

& TeCoA
& 76.70 & 95.70 & 64.86 & 91.30
& 43.50 & 75.50 & 36.62 & 71.90
& 50.36 & 83.28 & 37.76 & 73.61
& 23.74 & 55.26 & 17.93 & 47.92 \\

& PMG-AFT
& 80.00 & 96.60 & 65.82 & 91.32
& 55.70 & 85.10 & 43.88 & 78.74
& 52.26 & 85.18 & 37.89 & 73.50
& 32.16 & 65.18 & 22.14 & 54.96 \\

& FARE
& 84.90 & 98.60 & 70.26 & 93.96
& 61.60 & 89.40 & 49.20 & 82.98
& 54.12 & 86.56 & 40.93 & 76.20
& 34.52 & 67.66 & 25.78 & 59.86 \\

& TGA-ZSR
& 77.00 & 94.70 & 63.76 & 90.36
& 46.40 & 77.70 & 38.94 & 74.22
& 50.24 & 83.54 & 36.16 & 72.08
& 26.98 & 59.08 & 19.33 & 50.70 \\

& QT-AFT
& \underline{86.40} & \underline{98.80}
& \underline{70.72} & \underline{94.14}
& \underline{65.80} & \underline{91.00}
& \underline{51.90} & \underline{84.62}
& \underline{57.38} & \underline{87.46}
& \underline{41.93} & \underline{77.41}
& \underline{38.08} & \underline{72.24}
& \underline{27.47} & \underline{62.26} \\

& AGFT
& 79.60 & 96.90 & 66.42 & 91.82
& 58.10 & 86.40 & 45.64 & 79.76
& 53.34 & 85.74 & 38.40 & 74.24
& 33.42 & 67.26 & 22.80 & 56.42 \\

& AdvFLYP
& 80.50 & 97.10 & 65.92 & 91.52
& 59.20 & 87.10 & 46.96 & 80.84
& 52.32 & 84.64 & 38.08 & 73.53
& 33.50 & 67.70 & 24.16 & 57.67 \\

\rowcolor[RGB]{220,236,250}
\cellcolor{white}
\multirow{-9}{*}{\rotatebox{90}{\textbf{CoCa ViT-L/14}}}
& \textbf{TRAG (Ours)}
& \textbf{87.00} & \textbf{99.20}
& \textbf{71.40} & \textbf{94.66}
& \textbf{67.40} & \textbf{92.10}
& \textbf{54.02} & \textbf{86.10}
& \textbf{59.96} & \textbf{89.06}
& \textbf{42.98} & \textbf{78.07}
& \textbf{42.20} & \textbf{75.28}
& \textbf{29.26} & \textbf{64.37} \\

\bottomrule
\end{tabular}}
\end{table*}

\begin{table}[t]
\caption{\textbf{Text and multimodal attacks.} Flickr30k results for CLIP ViT-B/16 fine-tuned on CC3M. TR@m/IR@m average R@1 and R@10 in each direction.}
\vspace{-9pt}
\centering
\footnotesize
\setlength{\tabcolsep}{1pt}
\renewcommand{\arraystretch}{0.9}

\begin{tabularx}{0.927\columnwidth}{
    c*{6}{>{\centering\arraybackslash}X}
}
\toprule
\multirow{2}{*}[-0.6ex]{Method}
& \multicolumn{2}{c}{BERT-Attack}
& \multicolumn{2}{c}{Sep-Attack}
& \multicolumn{2}{c}{Co-Attack} \\
\cmidrule(lr){2-3}
\cmidrule(lr){4-5}
\cmidrule(lr){6-7}
& TR@m & IR@m & TR@m & IR@m & TR@m & IR@m \\
\midrule

TeCoA
& 41.85 & 34.74 & 26.05 & 21.79 & 51.65 & 38.76 \\

PMG-AFT
& 47.25 & 37.54 & 31.40 & 26.68 & 58.05 & 43.27 \\

FARE
& \underline{53.15} & 41.34 & 33.50 & 26.54
& 55.35 & 41.74 \\

TGA-ZSR
& 46.85 & 38.12 & 28.50 & 24.34
& 53.65 & 41.65 \\

QT-AFT
& 51.45 & \underline{42.50}
& 33.80 & \underline{28.69}
& 58.85 & \underline{45.46} \\

AGFT
& 49.80 & 38.32
& \underline{34.95} & 27.38
& \underline{59.00} & 43.77 \\

AdvFLYP
& 48.15 & 39.16 & 31.50 & 27.55
& 58.05 & 44.73 \\

\rowcolor[RGB]{220,236,250}
\textbf{TRAG (Ours)}
& \textbf{54.20} & \textbf{42.81}
& \textbf{35.40} & \textbf{29.46}
& \textbf{61.80} & \textbf{46.85} \\

\bottomrule
\end{tabularx}
\label{tab:flickr_attack_comparison}
\end{table}

\noindent\textbf{Different backbones.}
Table~\ref{tab:backbone_comparison} extends the comparison to ViT-B/16, ViT-L/14, and RN101.
TRAG again achieves the highest mean clean and robust accuracy for each backbone, supporting its applicability to both transformer and convolutional image encoders.
The ViT-L/14 model achieves 70.78\% clean and 49.54\% robust accuracy, improving on the best respective baselines by 1.72 and 2.76 points.

\subsection{Cross-Modal Retrieval}
\label{sec:retrieval}

We fine-tune CLIP ViT-B/16 and CoCa ViT-L/14 on CC3M~\cite{CC3M} and evaluate zero-shot retrieval on Flickr30k~\cite{Flickr30k} and COCO~\cite{COCO}.
We additionally include AdvFLYP~\cite{AdvFLYP} alongside the baselines in Sec.~\ref{sec:classification}.
Detailed settings are provided in the appendix.

Table~\ref{tab:retrieval_clean_robust} shows that TRAG achieves the best clean and robust recall among fine-tuned methods for both models, datasets, and retrieval directions.
For CoCa on COCO, robust TR@1 and IR@1 improve over QT-AFT by 4.12 and 1.79 percentage points, respectively.
These results extend the benefits to image--text fine-tuning.

\noindent\textbf{Text and multimodal attacks.}
Table~\ref{tab:flickr_attack_comparison} further evaluates CLIP on Flickr30k under text-only BERT-Attack~\cite{BERTAttack} and multimodal Sep-Attack and Co-Attack~\cite{CoAttack}.
TRAG leads on all six reported metrics, with gains of up to 2.80 percentage points.
Its observed benefits thus extend to text and joint image--text perturbations.

\subsection{Robustness Transfer to LVLMs}
\label{sec:lvlm_transfer}

We replace the image encoder of LLaVA-1.5-7B~\cite{LLaVA15} with the ViT-L/14 encoder fine-tuned on ImageNet in Sec.~\ref{sec:classification}, keeping the remaining modules unchanged.
Following FARE~\cite{FARE}, we evaluate captioning on COCO and Flickr30k and visual question answering on TextVQA~\cite{TextVQA} and VQAv2~\cite{VQAv2}.
Settings are provided in the appendix.

Table~\ref{tab:llava_robustness} shows that TRAG achieves the best robust results in seven of the eight task--attack combinations, while maintaining competitive clean performance.
On COCO, it reaches 73.47 CIDEr under Ensemble, exceeding the strongest baseline by 2.16 points.
These results extend TRAG's benefits to generative multimodal tasks without additional LVLM fine-tuning.

\begin{table*}[t]
\centering
\caption{\textbf{Robustness transfer to LLaVA-1.5-7B.} ImageNet-adapted ViT-L/14 encoders are evaluated using CIDEr for captioning and accuracy (\%) for VQA. Both attacks use an $\ell_\infty$ budget of $2/255$ in pixel space.}
\vspace{-9pt}

\label{tab:llava_robustness}

\resizebox{\linewidth}{!}{\setlength{\tabcolsep}{5pt}
\renewcommand{\arraystretch}{0.9}
\setlength{\heavyrulewidth}{1.0pt}
\setlength{\lightrulewidth}{0.7pt}

\begin{tabular}{c*{12}{c}}
\toprule
\multirow{2}{*}{\textbf{Method}}
& \multicolumn{3}{c}{\textbf{COCO}}
& \multicolumn{3}{c}{\textbf{Flickr30k}}
& \multicolumn{3}{c}{\textbf{TextVQA}}
& \multicolumn{3}{c}{\textbf{VQAv2}} \\
\cmidrule(lr){2-4}
\cmidrule(lr){5-7}
\cmidrule(lr){8-10}
\cmidrule(lr){11-13}
& Clean & APGD & Ensemble
& Clean & APGD & Ensemble
& Clean & APGD & Ensemble
& Clean & APGD & Ensemble \\
\midrule

TeCoA
& 101.94 & 53.76 & 40.41
& 59.07 & 32.23 & 19.05
& 24.48 & 12.80 & 13.26
& 65.60 & 39.76 & 40.92 \\

PMG-AFT
& 93.12 & 51.50 & 35.78
& 50.69 & 32.27 & 18.28
& 24.08 & 13.14 & 12.60
& 62.76 & 36.06 & 39.06 \\

FARE
& 115.67 & \underline{92.47} & 70.05
& \underline{72.88} & 54.34 & \underline{39.44}
& \underline{32.40} & \textbf{19.82} & \underline{20.90}
& 67.86 & \underline{48.14} & \underline{50.74} \\

TGA-ZSR
& 100.14 & 57.60 & 42.12
& 53.69 & 33.80 & 20.86
& 23.68 & 14.18 & 13.34
& 63.10 & 38.88 & 41.54 \\

QT-AFT
& \textbf{117.91} & 91.65 & \underline{71.31}
& \textbf{74.27} & \underline{54.49} & 37.92
& 32.26 & 19.44 & 19.82
& \underline{68.36} & 47.98 & 49.90 \\

AGFT
& 102.91 & 54.69 & 38.43
& 62.84 & 30.57 & 18.67
& 26.12 & 13.32 & 14.14
& 63.90 & 36.58 & 39.50 \\

\rowcolor[RGB]{220,236,250}
\textbf{TRAG (Ours)}
& \underline{116.15} & \textbf{93.77} & \textbf{73.47}
& 71.38 & \textbf{55.85} & \textbf{40.86}
& \textbf{32.68} & \underline{19.72} & \textbf{21.00}
& \textbf{69.26} & \textbf{48.74} & \textbf{51.14} \\

\bottomrule
\end{tabular}}
\end{table*}

\subsection{Ablation Studies and Analysis}
\label{sec:ablations_analysis}

\begin{table}[t]
\centering
\caption{\textbf{Pre-trained constraint ablations.} Mean accuracy (\%) over 16 datasets with ViT-B/32 at $2/255$. Representative quadratic mixtures and a direct FM+DM loss are compared under the same adversarial fine-tuning pipeline and task objective.}
\label{tab:guidance_ablation}
\footnotesize
\begin{tabularx}{\columnwidth}{l*{2}{>{\centering\arraybackslash}X}}
\toprule
Constraint & Clean & Robust \\
\midrule
FM & 58.69 & 34.44 \\
DM & 54.81 & 34.91 \\
Quadratic mix ($\lambda=5$) & 56.33 & 35.08 \\
Quadratic mix ($\lambda=10$) & 57.33 & 35.05 \\
FM + DM & 56.91 & 35.01 \\
\rowcolor[RGB]{220,236,250}
\textbf{TRAG (Ours)} & \textbf{59.60} & \textbf{35.09} \\
\bottomrule
\end{tabularx}
\end{table}

\begin{figure}[t]
\centering
\includegraphics[width=\columnwidth]{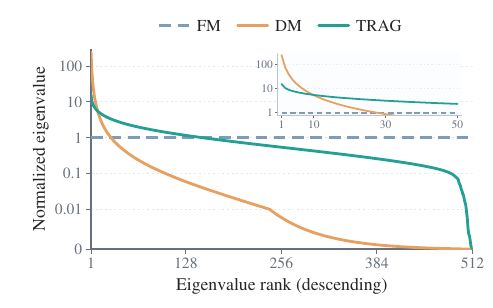}
\caption{\textbf{Directional weight spectra before fine-tuning.} Eigenvalues are sorted per image and averaged over 1,000 ImageNet images after normalizing each matrix to trace 512. The inset enlarges the first 50 ranks to show the leading directional weights.}
\label{fig:constraint_spectra}
\end{figure}

\begin{table}[t]
\centering
\caption{\textbf{Post-fine-tuning relationship preservation.} Means over 15 zero-shot datasets, with 64 images per dataset. C/A: clean/adversarial inputs. Relation error is scaled by $10^5$. Retention (\%) is measured on images classified correctly by pre-trained CLIP before adaptation and averaged across tasks.}
\label{tab:posterior_analysis}
\footnotesize
\setlength{\tabcolsep}{2pt}
\begin{tabularx}{\columnwidth}{l*{5}{>{\centering\arraybackslash}X}}
\toprule
\multirow{2}{*}{Guidance} & \multicolumn{2}{c}{Feature drift}
& \multicolumn{2}{c}{Relation error $\downarrow$} & Retention $\uparrow$ \\
\cmidrule(lr){2-3}\cmidrule(lr){4-5}
& C & A & C & A & A \\
\midrule
FM & 0.184 & 0.214 & 8.11 & 13.36 & 41.84 \\
DM & 0.332 & 0.353 & 7.71 & \textbf{11.07} & 41.97 \\
Quadratic mix & 0.202 & 0.229 & 7.47 & 11.90 & 42.00 \\
AGFT & 0.605 & 0.619 & 9.05 & 11.28 & 41.30 \\
FARE & 0.166 & 0.200 & 7.39 & 14.90 & 37.13 \\
\rowcolor[RGB]{220,236,250}
\textbf{TRAG (Ours)} & 0.249 & 0.277 & \textbf{6.68} & 11.22 & \textbf{42.66} \\
\bottomrule
\end{tabularx}
\end{table}

\noindent\textbf{Pre-trained constraints.}
We replace TRAG with FM, DM, and two combinations under the same fine-tuning pipeline.
Table~\ref{tab:guidance_ablation} compares a trace-normalized FM--DM quadratic mix and a direct FM+DM loss, with coefficients and search ranges detailed in the appendix.
The mixtures approach TRAG's robust accuracy but yield lower clean accuracy.
Against the quadratic mix with $\lambda=10$, TRAG gains 2.27 points in clean accuracy with nearly unchanged robust accuracy (35.09\% vs.\ 35.05\%).

\noindent\textbf{Directional weight spectra.}
Figure~\ref{fig:constraint_spectra} complements the constraint ablations with eigenvalue spectra before fine-tuning.
After trace normalization, FM assigns equal weights, DM concentrates its weights in relatively few directions, and TRAG distributes them more broadly while retaining nonuniform penalties.
These spectra illustrate constraint structure without directly predicting robustness.

\noindent\textbf{Relationship preservation.}
We examine 64 images per zero-shot dataset using clean inputs and adversarial candidates retained without fallback to the original image.
Relative to the pre-trained clean reference, we measure feature drift, the variance of class-similarity changes weighted by pre-trained probabilities, and retention of originally correct predictions.
The variance measures changes in pairwise class relationships; diagnostic details are provided in the appendix.
Table~\ref{tab:posterior_analysis} shows that TRAG allows greater feature drift than FM and the quadratic mix while preserving these relationships better, both before and after attack.
DM has slightly lower adversarial relationship error (11.07 vs.\ 11.22), while TRAG's prediction retention is numerically higher (42.66\% vs.\ 41.97\%); these differences remain within sampling uncertainty.
Relationship preservation does not imply robustness.

\section{Conclusion}
\label{sec:conclusion}

We presented TRAG, which guides adversarial fine-tuning of VLMs using local responses of the pre-trained text encoder.
Our analysis reveals complementary limitations of feature and prediction-distribution matching.
TRAG constructs anchor-level constraints from response directions and magnitudes, then aggregates them by image--anchor relevance to constrain visual representation drift.
Experiments demonstrate state-of-the-art clean and robust performance in zero-shot classification and cross-modal retrieval, with robustness benefits extending to LVLMs.
Ablations and posterior diagnostics support preserving pre-trained image--text relationships while allowing adaptation.
Local responses remain proxies for potential concept variations, and relationship preservation alone does not characterize robustness.
TRAG thus provides a practical way to use pre-trained text knowledge for robust visual adaptation without additional inference cost.

{
    \small
    \bibliographystyle{ieeenat_fullname}
    \bibliography{main}
}

\end{document}